\documentclass[a4paper,12pt]{article}
\usepackage[super,numbers,sort&compress]{natbib}
\usepackage{geometry}
\usepackage{graphicx}
\usepackage{amsmath}
\usepackage{amssymb}
\usepackage{multirow}
\usepackage{booktabs}
\usepackage{hyperref}
\newcommand{\bd}[1]{\mathbf{#1}}
\newcommand{\pathexpr}[3]{{#1}_{#2}^{(#3)}}
\newcommand{\ttt}[1]{\texttt{#1}}
\newcommand{\grad}[2]{\nabla_{\bd{#2}} #1}
\newcommand{\hess}[1]{\bd{H}_{\bd{#1}}}
\usepackage[ruled,vlined]{algorithm2e}
\usepackage{algorithmic}
\makeatletter 
\renewcommand\@biblabel[1]{#1.}
\makeatother

\begin{document}
	\title{\bf A general kernel boosting framework integrating pathways for predictive modeling based on genomic data}
	\author{\bf Li Zeng$^{1}$, Zhaolong Yu$^2$, Yiliang Zhang$^1$ and Hongyu Zhao$^{1,2,3,*}$}
	\date{
		$^1$ Department of Biostatistics, Yale University, New Haven, CT 06511, USA\\
	$^2$ Interdepartmental Program in Computational Biology and Bioinformatics, Yale University, New Haven, CT 06511, USA\\
	$^3$ Department of Genetics, Yale School of Medicine, New Haven, CT 06510, USA\\
	$^*$ Correspondence: hongyu.zhao@yale.edu
}
	\maketitle
	\begin{abstract}
Predictive modeling based on genomic data has gained popularity in biomedical research and clinical practice by allowing researchers and clinicians to identify biomarkers and tailor treatment decisions more efficiently. Analysis incorporating pathway information can boost discovery power and better connect new findings with biological mechanisms. In this article, we propose a general framework, Pathway-based Kernel Boosting (PKB), which incorporates clinical information and prior knowledge about pathways for prediction of binary, continuous and survival outcomes. We introduce appropriate loss functions and optimization procedures for different outcome types. Our prediction algorithm incorporates pathway knowledge by constructing kernel function spaces from the pathways and use them as base learners in the boosting procedure. Through extensive simulations and case studies in drug response and cancer survival datasets, we demonstrate that PKB can substantially outperform other competing methods, better identify biological pathways related to drug response and patient survival, and provide novel insights into cancer pathogenesis and treatment response.
		\end{abstract}

		\begin{center}
			\end{center}

	\section{Introduction}

High-throughput genomic technologies have proved powerful to advance our understanding of the associations between genes and clinical phenotypes of interest. \cite{reuter2015high} The analyses of these data have provided valuable insights into cancer mechanisms, pathogenesis and treatment response. \cite{kim2013gwas, sanchez2018oncogenic, liu2018integrated}  Because signals of a single gene in clinical phenotypes can be  weak and one gene is often involved in multiple biological processes, gene level results are often unstable and lack interpretability. Pathway-based analysis offers a more informative and robust alternative, since it can aggregate signals from individual genes and its results are more easily interpretable. As a result, pathway-based methods have gained popularity and led to successful identifications of many disease and/or treatment relevant pathways with better power and interpretability in recent years.\cite{subramanian2005gene, ramanan2012pathway} It has been found that the onset and progression of a disease is usually affected by many pathways.\cite{shou2004mechanisms, shtivelman2014molecular, berk2009neuroprogression} However, most of the exsiting pathway-based methods estimate the effects on phenotypes or evaluate the statistical significance for each pathway separately, and thus cannot quantify the effect of a given pathway in addition to other pathways.\cite{subramanian2005gene,liu2007semiparametric,wu2011rare} Therefore, there is a need for methods that integrate all the pathways in the model at the same time.

Nonparametric Pathway-based Regression (NPR)\cite{wei2007nonparametric} and Group Additive Regression (GAR)\cite{luan2007group} are two of the early attempts to jointly model all pathways. They extended the Gradient Descent Boosting (GDB)\cite{friedman2001greedy} framework to minimize the empirical loss function of different types of dependent variables. GDB is a functional gradient descent algorithm that minimizes the empirical loss function. At each iteration of the algorithm, an increment function in the space of base learners is selected to achieve the steepest descent to the loss function, and added to the additive model. NPR uses regression trees while GAR uses linear models to construct base learners from different pathways. However, due to the linearity assumption of GAR, it lacks the ability to capture complex interactions among genes in the same pathway. Using regression trees as base learners, NPR can model interactions but there is no regularization in the gradient descent step, which can lead to selection bias for large pathways.

%It is shown that gene-gene interactions have stronger effects on phenotypes when the genes belong to the same pathway or regulatory network.\cite{carlson2004mapping} To capture these interactions, multiple kernel methods have been commonly used and achieved state-of-the-art performance in predictions of various outcomes.\cite{gonen2014drug,costello2014community, aiolli2015easymkl,friedrichs2017pathway, manica2019pimkl} In these methods, one kernel is assigned to each group of predictors and a meta-kernel is computed as a weighted sum of the individual kernels. Kernel weights are estimated through optimization and considered as a measure of pathway importance. 

In a previous work,\cite{zeng2019pathway} we proposed a pathway-based kernel boosting (PKB) algorithm that can utilize patients' gene expression data and prior pathway knowledge from existing databases to perform binary classification. We used the second order approximation of the loss function instead of the first order approximation which is used in the classic gradient descent boosting method, to allow for deeper descent at each step. A kernel function space was constructed from each pathway, and the union of the kernel function spaces was used as the base learner space. From theories of the Reproducing Kernel Hilbert Space (RKHS)\citep{friedman2001elements}, this base learner space is flexible enough to capture complex gene-gene interactions. In order to prevent overfitting, we introduced two types of regularizations ($L_1$ and $L_2$) for selection of base learners at each iteration. However, two drawbacks of the model limit the usefulness of PKB in cancer data analysis. First, typical cancer genomic study datasets provide both patients' genomic features as well as clinical features, but only genomic features are considered as predictors in this method, which cannot capitalize on the rich clinical information that is potentially predictive. Second, PKB can only model binary outcome variables, which excludes the analysis for other types of interesting clinical outcomes, such as drug response, disease free survival, and overall survival. 
	
In this article, we propose an extended PKB framework. It enables the inclusion of clinical features as predictors by adding a linear model component to the base learner spaces. It also unifies classification, regression, and survival analysis under the same boosting procedure, which handles different types of outcome variables by specifying different loss functions. We provide details of the proposed framework in \textbf{Section \ref{methods}}. Through extensive simulations in \textbf{Section \ref{simu}}, we demonstrate that PKB substantially outperforms competing methods including LASSO,\cite{tibshirani1996regression} Ridge Regression,\cite{hoerl1970ridge} and ElasticNet\cite{zou2005regularization} in predicting continuous outcomes, and Glmnet,\cite{simon2011regularization} RandomSurvivalForest,\cite{ishwaran2008random} and CoxBoost\cite{binder2013coxboost} in predicting survival outcomes. For applications, we evaluate the  performances of our proposed method on two databases:  the Cancer Cell Line Encyclopedia (CCLE) \cite{barretina2012cancer} for cell line drug response prediction, and The Cancer Genome Atlas (TCGA) \cite{tomczak2015tcga} for patient survival prediction in \textbf{Section \ref{results}}.

	\section{Materials and methods}\label{methods}
	Suppose our observed data are collected from $N$ subjects. For subject $i$, we use a $p$ dimensional vector $\bd{x}_i = (x_{i1}, x_{i2}, \ldots, x_{ip})$ to denote his/her normalized gene expression profile. Similarly, the gene expression levels of a given pathway $m$ with $p_m$ genes can be represented by $\pathexpr{\bd{x}}{i}{m} = (\pathexpr{x}{i1}{m},\pathexpr{x}{i2}{m}, \ldots, \pathexpr{x}{ip_m}{m})$, which is a sub-vector of $\bd{x}_i$. We use $\bd{z}_i = (z_{i1}, z_{i2}, \ldots, z_{iq})$ to represent available clinical features for this subject, such as sex, age, and cancer type.
	
	For classification, each subject has an observed class label $y_i \in \{1,-1\}$. The probability of being class 1 is modeled as 
	$$p(y = 1| \bd{x}, \bd{z}) = \frac{\exp[F(\bd{x}, \bd{z})]}{1 + \exp[F(\bd{x}, \bd{z})]},$$
	where $F(\bd{x}, \bd{z})$ is the log odds function of the classification model. Maximizing the likelihood is equivalent to minimizing the following log loss function:\cite{zeng2019pathway}
	$$l(y , F(\bd{x}, \bd{z}) ) = \log(1+\exp[-y F(\bd{x}, \bd{z})]).$$
	
	In the regression model, we observe continuous outcome $y$ for each subject. The loss function for regression is the commonly used squared error:
	$$l(y , F(\bd{x}, \bd{z}) ) = ( y - F(\bd{x}, \bd{z}))^2$$
	
	In the survival model, outcome for each subject is a bivariate tuple $y = (t, \delta)$, where $t$ is the survival time or censoring time, and $\delta$ is an indicator of the endpoint event, such as disease relapse or death in most cancer studies. When $\delta = 1$, $t$ is the actual survival time and when $\delta = 0$, $t$ is the censoring time. We build the survival model following the Cox regression's assumption on the hazard function, but replacing the linear component with a nonlinear risk score function $F(\bd{x}, \bd{z})$: \cite{li2005boosting}

	\begin{equation}
	\label{eqn:hazard}
	h(t, F(\bd{x},\bd{z}) ) = h_0(t)\exp[F(\bd{x},\bd{z})]
	\end{equation}
	where $h_0(t)$ is an unknown baseline hazard function. Using the partial likelihood function allows us to circumvent the inference of $h_0(x)$ and directly estimate $F(\bd{x}, \bd{z})$. We use the negative log partial likelihood\cite{li2005boosting} as the loss function in the survival model:
	$$l(y, F(\bd{x}, \bd{z}) ) = -\delta \left\lbrace F(\bd{x}, \bd{z}) - \log\left( \sum^{N}_{j=1}1_{\left\lbrace t_j \geq t\right\rbrace }\exp[F(\bd{x}, \bd{z})]\right)  \right\rbrace.$$
	
	The goal of PKB is to estimate $F(\bd{x}, \bd{z})$ non-parametrically through a boosting procedure. $F(\bd{x}, \bd{z})$ has different interpretations in different loss functions. We will refer to it as the score function for the rest of the article.
	\subsection{Base learner space}
	Results from theories of RKHS\cite{friedman2001elements} have shown that kernel functions can capture complex interactions among input features. It was used in Zeng et al.\citep{zeng2019pathway} to construct base learner spaces for classification purpose. We extend this formulation by incorporating clinical features as a linear component in the base learner space. For pathway $m$, the base learner space takes the following form:
	\begin{equation}
	\label{eqn:G}
	\mathcal{G}_m = \{  f(\bd{x}, \bd{z}) = \sum_{i=1}^N K_m(\pathexpr{\bd{x}}{i}{m}, \pathexpr{\bd{x} }{}{m}) \beta_i + \sum_{j=1}^q z_j \gamma_j : \mathbf{\beta} \in R^{N}, \mathbf{\gamma} \in R^{q} \},
	\end{equation}
	where $K_m(\cdot,\cdot)$ is a kernel function that defines the similarity between two subjects using only genes in the $m$th pathway. Each element in the space is composed of two parts: a nonlinear component to model gene expression effect and a linear component to model clinical features' effects. The overall base learner space $\mathcal{G}$ is the union of individual learner spaces from each pathway: $\mathcal{G} = \bigcup_{m=1}^M\mathcal{G}_m$.
	
	Assume that there are $M$ pathways considered in our model. Since genes within the same pathway likely have stronger interactions than genes in different pathways, we assume additive effects across pathways and focus on capturing gene interactions within pathways:
	\begin{equation*}
	F(\bd{x}, \bd{z}) = \sum_{m=1}^M H_m(\bd{x}^{(m)}, \bd{z}),
	\end{equation*}
	where each $H_m$ is a function that only depends on the expression level of genes in the $m$th pathway and the clinical features. $H_m$ belongs to the RKHS of the $m$th pathway $\mathcal{G}_m $. Due to the additive nature of this model, it only captures gene interactions within each pathway but not across pathways. 
	
	Given a specific response type and the corresponding loss function, the empirical loss is defined as the mean loss evaluated at each sample point:
	$$L(\bd{y},\bd{F}) = \frac{1}{N} \sum_{i=1}^N l(y_i,F(\bd{x}_i,\bd{z}_i)),$$
	where $\bd{F} = (F(\bd{x}_1, \bd{z}_1), F(\bd{x}_2, \bd{z}_2), \ldots, F(\bd{x}_N, \bd{z}_N))$. For the rest of this article, we will also use the boldface font of a function to represent the vector of the function evaluated at each observed sample point.
	
	\subsection{Identification of the optimal increment function } \label{sec:identify}
	Boosting is an iterative functional descent procedure to minimize the empirical loss function. Assume that after iteration $t$, the estimated target function is $F_t(\bd{x}, \bd{z})$. In the next iteration, we want to identify an ``optimal" increment function $f \in \mathcal{G}$, which shrinks the current empirical loss as much as possible, and then add it to $F_t(\bd{x}, \bd{z})$ to yield $F_{t+1}(\bd{x}, \bd{z})$. 
	
 We approximate the loss function $L(\bd{y},\bd{F_{t+1}})$ by its second order Taylor expansion around $\bd{F_t}$:
	\begin{eqnarray*}
		L(\bd{y}, \bd{F_t} + \bd{f_t}) &\approx& L(\bd{y}, \bd{F_t}) +  (\grad{L}{F_t})^T \bd{f_t} + \frac{1}{2}\bd{f_t}^T \hess{F_t} \bd{f_t} \\
		& = & \frac{1}{2} (\bd{f_t} + \hess{F_t}^{-1} \grad{L}{F_t})^T \hess{F_t} (\bd{f_t} + \hess{F_t}^{-1} \grad{L}{F_t}) + \mbox{const},
	\end{eqnarray*}
where $\bd{f_t}$ is the increment direction at iteration $t$. $\nabla_{\bd{F_t}} L$ and $\bd{H}_{\bd{F_t}}$ are the gradient and Hessian matrix of $L(\bd{y}, \bd{F_t})$ with respect to $\bd{F_t}$, respectively, and const includes all the terms that do not involve $\bd{f_t}$. Note that this approximation is accurate in the case of regression, since the regression loss is quadratic itself. Closed form expressions of $\grad{L}{F_t}$ and $\hess{F_t}$ for different problems are given in \textbf{Appendix A} of the \textbf{Supplementary Materials}. Direct minimization for this loss approximation in $\mathcal{G}$, however, may lead to overfitting, due to the high flexibility of the base learner space. It is necessary to apply penalties in the selection of the increment function  $f_t$. Here we propose the following regularized loss as the working loss function in PKB:
$$L_{R}(\bd{f_t}) = \frac{1}{2} (\bd{f_t} + \hess{F_t}^{-1} \grad{L}{F_t})^T \hess{F_t} (\bd{f_t} + \hess{F_t}^{-1} \grad{L}{F_t}) + \lambda \Omega({f_t}),$$
where $\Omega({f_t})$ is the penalty term. Since ${f_t}$ takes the functional form as presented in \textbf{Equation (\ref{eqn:G})}, it is natural to consider the  $L_1$ and $L_2$ norms of $\beta_t$ penalty: $\Omega({f_t}) = \| \beta_t \|_1$ or $\| \beta_t \|_2^2$. Such a penalized boosting step has been employed in several methods. \cite{johnson2014learning} Intuitively, the regularized loss function would prefer simple solutions that also fit the observed data well, which usually leads to better generalization capability to unseen data.

Optimizing $L_R(\bd{f_t})$ in $\mathcal{G}_m$ is equivalent to solving
\begin{equation}
\label{eqn:regloss}
\min_{\beta_t, \gamma_t} \frac{1}{2} (K_m \beta_t + Z \gamma_t + \hess{F_t}^{-1} \grad{L}{F_t})^T \hess{F_t} (K_m \beta_t + Z \gamma_t + \hess{F_t}^{-1} \grad{L}{F_t}) + \lambda \Omega({f_t}),
\end{equation}
where $K_m$ is the $N \times N$ kernel matrix calculated using gene expressions in the $m$th pathway, and $Z$ is the $N \times q$ clinical feature matrix. It can be proved that, by applying the following transformation:
\begin{eqnarray*}
	\tilde{\eta}_t &=& \frac{1}{\sqrt{2}}\hess{F_t}^{\frac{1}{2}}(I_N - Z(Z^T \hess{F_t} Z)^{-1}Z^T \hess{F_t})\hess{F_t}^{-1}\grad{L}{F_t}\\
	\tilde{K}_m &=& \frac{1}{\sqrt{2}}\hess{F_t}^{\frac{1}{2}}(I_N - Z(Z^T\hess{F_t}Z)^{-1}Z^T\hess{F_t})K_m,
\end{eqnarray*}
solving $\beta_t$ in \textbf{Equation (\ref{eqn:regloss})} is reduced to:
\begin{equation}
\label{eqn:reduced}
\min_{\beta_t} \|\tilde{\eta}_t + \tilde{K}_m \beta_t \|_2^2 + \lambda \Omega({f_t}).
\end{equation}
Proof is provided in \textbf{Appendix B} in the \textbf{Supplementary Materials}. \textbf{Equation (\ref{eqn:reduced})} becomes the LASSO problem when we use the $L_1$ penalty, and the Ridge Regression when we use $L_2$ penalty. Both can be efficiently solved with existing solvers. After solving $\beta_t$, we subsequently obtain the solution of $\gamma_t$ as:
$$\gamma_t = -(Z^T\hess{F_t}Z)^{-1}Z^T\hess{F_t}(K_m\beta_t + \hess{F_t}^{-1}\grad{L}{F_t}).$$

With $\beta_t$ and $\gamma_t$ solved, we obtain the best increment direction $f_t$.
%Given the best increment direction $\bd{f_t}$ at iteration $t$, we find the deepest step length by minimizing over the original loss function:
%\begin{equation*}
%d_t = \text{arg}\min_{d_t\in R^+} L\left(y, \bd{F_t}+d_t\bd{f_t}\right),
%\end{equation*}
%and update the target function to $\bd{F_{t+1}}\left(\bd{x}, \bd{z}\right) = \bd{F_t}\left(\bd{x}, \bd{z}\right) + \nu d_t\bd{f_t}$, where $\nu$ is a learning rate parameter. The above fitting procedure is repeated until a certain pre-specified number of iterations is reached.

\subsection{The PKB algorithm}
In this section we propose the PKB algorithm that solves classification, regression, and survival analysis in a unified framework in \textbf{Algorithm \ref{algorithm: pbk}}. $F_t(\bd{x}, \bd{z})$ is the estimated score function at iteration $t$, and $F_T(\bd{x}, \bd{z})$ is the final score function estimation.

\begin{algorithm}\label{algorithm: pbk}
\caption{Unified PKB framework}
\SetAlgoLined
\textbf{Input}: normalized gene expreesion profile with pathway information $\bf{x}$; clinical features $\bf{z}$; outcome $y$; loss function $l$; learning rate parameter $\nu \in (0,1)$; iterations number $T$.\\
\textbf{Initialization}: Initialize $F_0(\bd{x}, \bd{z})$ as a constant that minimizes the empirical loss
$$F_0(\bd{x}, \bd{z}) = \arg\min_{c} \frac{1}{N} \sum_{i=1}^N l(y_i, c).$$
In the case of survival model, we set $F_0(\bd{x}, \bd{z}) = 0$, because the partial likelihood of Cox model is not affected by constants. \\

\While{$T >0$}{
	\begin{enumerate}
		\item \textbf{Identify the optimal increment function $f_t$}. \\
		Calculate the gradient $\grad{L}{F_t}$ and Hessian matrix $\hess{F_t}$. For each pathway $m$, solve the optimal $\hat{f}_m \in \mathcal{G}_m$ and corresponding $\beta, \gamma$ following the steps in \textbf{Section \ref{sec:identify}}. The optimal increment function $f_t$ is the $\hat{f}_m$ which yields the smallest $L_R(\bd{f_t})$.
		\item \textbf{Line search and update $F_t$}\\
		Decide the step length. First perform a line search
		$$ d_t = \arg\min_{d \in R^{+}} L(\bd{y}, \bd{F_t} + d\bd{f_t}).$$
		Then apply a shrinkage on the step size using the learning rate parameter $\nu$, and update the estimate of the score function:
		$$F_{t+1}(\bd{x}, \bd{z}) = F_t(\bd{x}, \bd{z}) + \nu d_t \bd{f_t}(\bd{x}, \bd{z}).$$
	\end{enumerate}
	$T\gets T-1$
}
\end{algorithm}

The learning rate parameter $\nu$ takes value in $(0, 1)$, usually smaller than 0.1. Our experience shows that the combination of the line search technique and the learning rate parameter makes the model fitting procedure more stable, and less prone to overfitting.

The choice of $T$ is critical to prediction accuracy on test data. Choosing a T-value that is  too small or too large can lead to underfitting and overfitting, respectively. We employ a cross-validation procedure to determine the number of iterations for $T$. We split the dataset into three folds, and simultaneously initiate three runs of \textbf{Algorithm \ref{algorithm: pbk}}, each using two folds as training data and the other fold as testing data. After each iteration, a cross-validated loss is calculated by averaging the three losses. We keep track of the minimum cross-validated loss. If the minimum loss value does not change in 50 iterations, we end the cross-validation process, and use the iteration with the minimum loss as the number of iterations T.

Both the $L_1$ and $L_2$ boosting algorithms require the specification of the penalty parameter $\lambda$, which controls step length (the norm of fitted $\beta$) in each iteration for $L_1$ and $L_2$ and additionally controls solution sparsity in $L_1$ case. Poor choices of $\lambda$ can result in big leaps or slow descent speed. To address this issue, we also incorporate an optional automated procedure to choose the value of $\lambda$ in PKB. Computational details of the procedure are provided in Section 2 of the Supplementary Materials of Zeng et al.\cite{zeng2019pathway}

After fitting the PKB model, the final score function estimate takes the form
$$F_T(\bd{x}, \bd{z}) = \sum_{m=1}^M \sum_{i=1}^N K_m(\pathexpr{\bd{x}}{i}{m},\pathexpr{\bd{x}}{}{m})\pathexpr{\beta}{i}{m} + \bd{z}^T \gamma,$$
where $\pathexpr{\beta}{}{m}$ is the coefficient vector for pathway $m$. The values of $\pathexpr{\beta}{}{m}$ can be used to evaluate the significance of pathways in the score function. We propose to use the $L_2$ norm, $w_m = \|\beta^{(m)}\|_2$, as weights for pathways. Note that $w_m$ is non-zero only if the pathway is selected at least once during model fitting. From our experience, in applications with a large number of input pathways, many pathways will end up with zero weights.

\section{Simulation Study}\label{simu}
In this section, we apply PKB to a variety of simulation datasets, and demonstrate that it can yield better prediction accuracy compared to competing methods, as well as correctly identify informative pathways for regression and survival models. 

We design the following three models for the underlying true score functions $F(\bd{x},\bd{z})$: 
	 \\\textbf{Model 1},
	\\\centerline{$F(\bd{x},\bd{z}) = 3z_1 - 4z_2 + 3z_3 + 2\pathexpr{x}{1}{1}+3\pathexpr{x}{2}{1}+ 
	3\exp(0.5\pathexpr{x}{1}{2} + 0.5\pathexpr{x}{2}{2}) + 4\pathexpr{x}{1}{3}\pathexpr{x}{2}{3}$; }
	\\\textbf{Model 2},
	\\\centerline{$F(\bd{x},\bd{z}) = z_1 - 3z_2 + 3z_3 - z_4 + 6\sin(0.5\pathexpr{x}{1}{1} + 0.5\pathexpr{x}{2}{1}) + 2\log(|{\pathexpr{x}{1}{2}}^3 - {\pathexpr{x}{2}{2}}^3|) + 2({\pathexpr{x}{1}{3}}^2-{\pathexpr{x}{2}{3}}^2)$; }
	\\\textbf{Model 3},
	\\\centerline{$F(\bd{x},\bd{z}) = z_1 + z_3 + 2\sum_{m=1}^{8} \|\pathexpr{\bd{x}}{}{m}\|_2.$}

In the above equations, $\pathexpr{x}{i}{m}$ represents the  expression level for the $i$th gene in pathway $m$. The three models involve a wide variety of functional forms of pathway effects, including linear, polynomial, exponential, logarithm, and sine effects. We assume the effects from clinical variables are linear. 

Under each model, we simulate two datasets, with 20 and 50 pathways, respectively. Note that in \textbf{Models 1} and \textbf{2}, only the first three pathways are informative, and in \textbf{Model 3}, the first eight pathways are informative. The predictive signals in the datasets with 50 pathways are sparser than the 20-pathway datasets. For each pathway, we simulate expression levels for five genes using Gaussian distribution. In each dataset, we also generate five clinical features: two binary features generated from Bernoulli distribution, and three continuous features generated from Normal distribution. The number of predictive clinical variables is three, four, and two for the three models, respectively. The sample size for all datasets is 300.

The outcome values $\bd{y}$ for the regression and survival models are generated under different mechanisms. For the regression model, we add a Gaussian noise to the $F(\bd{x}, \bd{z})$ values to generate $\bd{y}$. The variance of the Gaussian noise is set to one-fifth of the $F(\bd{x}, \bd{z})$ values' variance.

For simulation of survival outcomes, we assume a Weibull baseline hazard $h_0(t) = \kappa \rho t^{\rho-1}$, and cumulative hazard function $H_0(t) = \kappa t^{\rho}$, where $\kappa$ and $\rho$ are the scale and shape parameters, respectively. Suppose that the score $F(\bd{x},\bd{z})$ is calculated for one sample. A corresponding survival time can be generated from
$$t = \left( - \frac{\log(U)}{\kappa \exp[F(\bd{x},\bd{z})]} \right)^{\frac{1}{\rho}},$$
where $U$ is randomly drawn from $\mbox{Uniform}(0,1)$ distribution. \citep{bender2005generating} The values of $\kappa$ and $\rho$ are chosen such that the median survival time is 20 months, which is on the same scale as the median survival times of many cancer types. We then randomly draw 20\% of the samples for censoring, and the censoring times are drawn from a uniform distribution between zero and the generated survival times.

When evaluating prediction performance, we use mean square error (MSE) for regression, and C-index for the survival model.\citep{harrell1982evaluating} C-index is commonly used in assessing survival prediction accuracy. In general, C-index looks at all possible pairs of samples, and calculates the ratio of the pairs where the predicted risk scores are concordant with the observed survival times. If the predicted risk score is not informative, C-index would be close to 0.5. On the contrary, if the model perfectly predicts risk score, C-index would be 1. More details regarding the calculation of C-index can be found in \textbf{Appendix C} in the \textbf{Supplementary Materials}. For each dataset, we perform ten runs of the PKB algorithm. In each run, we use two-thirds of the samples as training data, and assess prediction performance on the remaining samples. 

\subsection{Simulation results for the regression model}
We compared the performance of the PKB regression model to several existing methods, including LASSO,\cite{tibshirani1996regression} Ridge Regression,\cite{hoerl1970ridge} and ElasticNet, \citep{zou2005regularization} which are linear models, and RandomForest,\citep{breiman2001random} Gradient Boosting Regression (GBR), \citep{friedman2001greedy} and Support Vector Regression (SVR),\citep{smola2004tutorial} which are nonlinear models. We extensively tuned the parameters for all methods, and the configuration details are provided in \textbf{Appendix D} in the \textbf{Supplementary Materials}. \textbf{Table \ref{tab:simu_reg}} compares the average MSE on test data over 10 runs. Standard deviations of the MSEs can be found in \textbf{Appendix H} in the \textbf{Supplementary Materials}.

\begin{figure}[htp]
		\centering
		\includegraphics[width=1.07\textwidth]{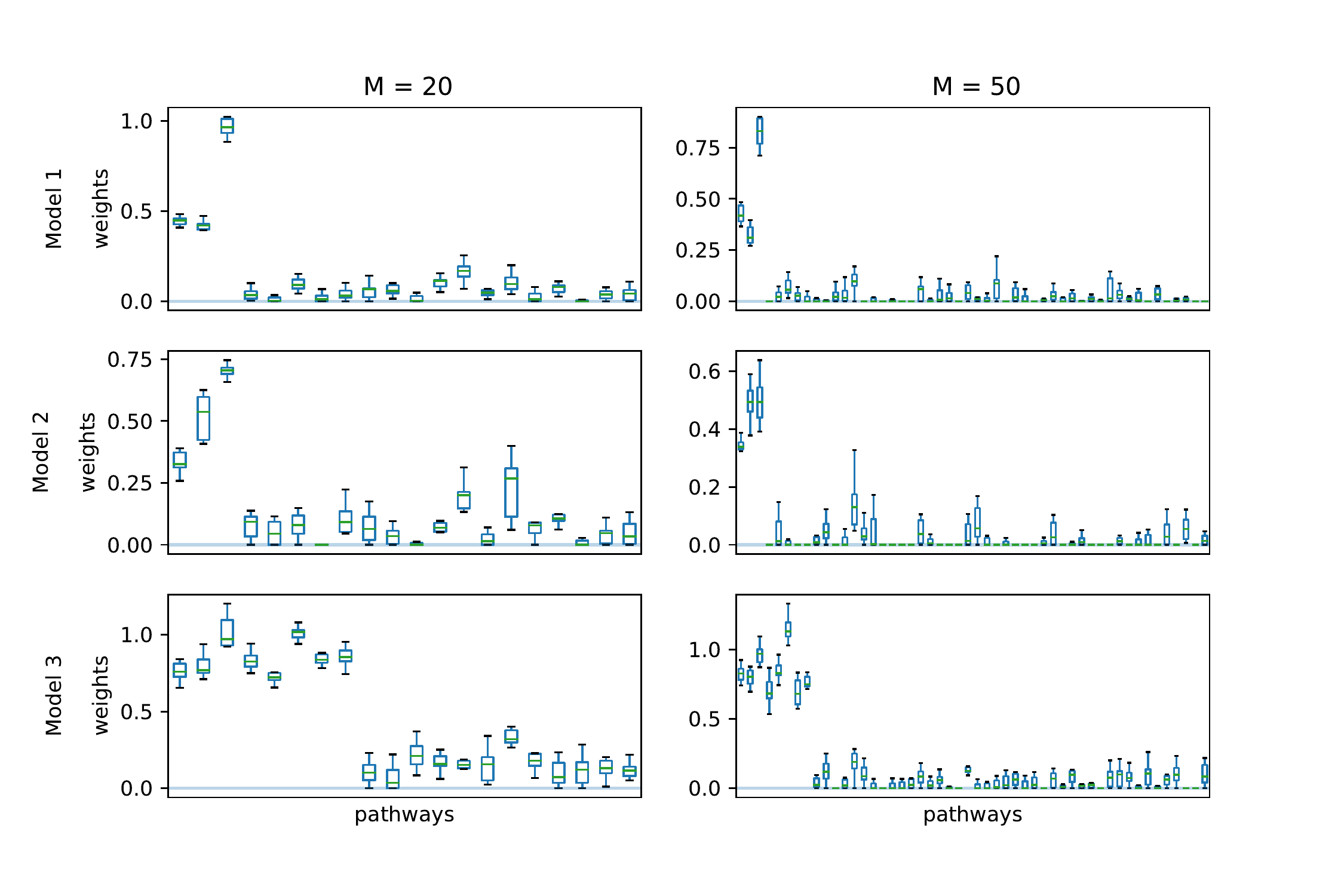}
		\caption{Boxplots for pathway weights in the regression simulations. Each box represents the weights distribution for one pathway over ten PKB runs.}
		\label{fig:reg_weights}
	\end{figure}

In all simulation scenarios, the two PKB algorithms, PKB-$L_1$ (with $L_1$ model complexity penalty) and PKB-$L_2$ (with $L_2$ model complexity penalty), yielded significantly better prediction accuracy than competing methods. Among the competing methods, the sparse linear models had better accuracy in Models 1 and 2, where only three pathways are informative. Since there are eight pathways relevant to the outcome in Model 3, the nonlinear genomic signal becomes stronger, thus the nonlinear methods produced equal or better accuracy than the linear methods. We also assessed the ability of PKB to properly weigh the informative pathways. For each pathway, we calculated its weights in the final score function over the ten runs. The distributions of the weights are presented in \textbf{Figure \ref{fig:reg_weights}}. In all simulation scenarios, the PKB algorithm gave relevant pathways significantly higher weights than the other pathways. Note that PKB shrank the weights of some, but not all the noise pathways to zero. This is expected, because as the boosting procedure continues, the predictive effect from true informative pathways becomes weaker, and the noise pathways may occasionally result in the smallest loss and subsequently be selected in certain iterations.

\subsection{Simulation results for the survival model}
In the survival analysis simulations, we compared our method with Glmnet,\citep{simon2011regularization} RandomSurvivalForest, \citep{ishwaran2008random} and CoxBoost. \citep{binder2013coxboost} Glmnet is an extension of the Cox regression model with penalties, and is commonly used in the analysis of survival data with high dimensional predictors. RandomSurvivalForest and CoxBoost are also extensions of Random Forest and CoxBoost, respectively,  to perform survival analysis. Predictive performance was evaluated using C-index. The average C-index for each method over the ten runs is presented in \textbf{Table \ref{tab:simu_surv}}, and the standard deviations are available in \textbf{Appendix H} in the \textbf{Supplementary Materials}. 

Both PKB methods significantly outperformed the competing methods in all  simulation scenarios. We also examined the pathway weights, similar as in the regression simulations. The results suggest that the PKB survival model could effectively identify informative pathways and weigh them properly in the score function. The weights distribution figure has a similar pattern to \textbf{Figure \ref{fig:reg_weights}} from the regression simulations and we leave it in \textbf{Appendix F} in the \textbf{Supplementary Materials}.

\section{Applications}\label{results}
In order to examine the performance of our proposed model on real datasets, we applied PKB, along with all the competing methods, to two cancer-related databases: CCLE\citep{barretina2012cancer} for regression analysis, and TCGA for survival analysis. 

We followed the same procedure as that in simulation studies to assess the performances of the methods. In the applications of PKB, pathway annotation databases, including the Kyoto Encyclopedia of Genes and Genomes (KEGG), \citep{kanehisa2000kegg} Biocarta, \citep{nishimura2001biocarta} and, Gene Ontology Biological Process (GO-BP) \citep{ashburner2000gene, gene2016expansion}, were used as sources of pathway information. Details about the choices of model parameters can be found in \textbf{Appendix D} in the \textbf{Supplementary Materials}.

\subsection{The CCLE drug response prediction}

The CCLE is a rich database containing cancer cell line responses to anti-cancer compounds, and involves cell lines from over 20 cancer types and 24 anticancer compounds with various targets. A compilation of RNA-seq gene expression data is available for about 1000 cell lines, which enables the analysis of the association between genes and drug responses. 

Drug response is measured by the IC50 value in the CCLE database, which is defined as the concentration needed for the compound to kill 50$\%$ of the tumor cells in the cell culture. \citep{barretina2012cancer} In our prediction implementation, the log-transformed IC50 value was used as the outcome variable for regression. For clinical predictors, we considered cancer types and the gender of the cell line provider. We applied our method to predict responses for six compounds (named after their corresponding targets) that have sufficient sample sizes: EGFR, HDAC, MEK, RAF, TOP, and TUBB1. 

\textbf{Table \ref{tab:ccle_mean}} demonstrates the cross-validated prediction MSEs from PKB and all competing methods, with the top two methods marked in bold. In five out of the six datasets, at least one of the PKB methods appeared in the top two methods. In the MEK dataset, both PKB-$L_1$ and PKB-$L_2$ were ranked as the top two methods, and there was substantial difference in MSE comparing our methods with the competing methods (\textbf{Appendix H} in the \textbf{Supplementary Materials}). Among pathways that have been identified as important for the drug response, the KEGG Asthma pathway is informative for drug responses to MEK inhibitors and RAF inhibitors.

\subsection{The TCGA cancer patient survival prediction}

In order to assess the predictive performance of PKB on real survival data, we applied the PKB survival model to seven cancer study datasets, including (by primary tumor sites): brain, head/neck, skin, lung, kidney, stomach, and bladder. The datasets were selected using criteria such as large sample size with gene expression data and low censoring ratio. The TCGA datasets include more complete clinical information compared to the cell line datasets in CCLE. The clinical features we used generally included patient gender, age, tumor subtype, site/laterality, and stage, which were available in almost all datasets and had low missing rates. 

The cross-validated prediction C-indices from all methods are presented in \textbf{Table \ref{tab:tcga_mean}}. The PKB methods were the top two methods for accuracy in five out of the seven datasets, and there was substantial difference in C-index between PKB methods and the competing methods (\textbf{Appendix H} in the \textbf{Supplementary Materials}). In the remaining two cases, PKB also yielded performances (C-index difference $< 0.01$, not significant) comparable to the top two methods. In the brain and kidney datasets, PKB was most successful, which outperformed the third best method by 0.26 and 0.15 in terms of C-index, respectively. For these two cancer patient survival prediction results, we also backtracked the biological pathways, which are closely related to the outcome. For the KEGG pathway, the neuroactive ligand receptor interaction turned out to be the most significant for brain tumor patients' survival, and for the GO pathway, homophilic cell adhesion via plasma membrane seemed to be the most important pathway according to the boosting procedure.

We further performed pathway enrichment analysis, and examined the p-values of the pathways considered significant by PKB. (If a pathway took positive weights in at least four out of ten runs of PKB, it was considered significant.) The enrichment analysis was conducted on each dataset following the procedure in \textbf{Appendix E} in the \textbf{Supplementary Materials}.
The enrichment analysis results for the brain, kidney, and lung cancer datasets are presented in \textbf{Figure \ref{fig:gsea}}. Results for other datasets are available in \textbf{Appendix E} in the \textbf{Supplementary Materials}. We observe different enrichment patterns from different datasets. In the lung dataset (bottom panel of \textbf{Figure \ref{fig:gsea}}), the PKB significant pathways are highly concentrated to the left, meaning that they are also considered significant in the enrichment analysis. In the brain and kidney cancer, the PKB pathways are more spread out: some of them appear at the top, but many others are not considered significant by the enrichment analysis. This is not surprising, because enrichment analysis looks at marginal associations between pathways and outcomes, while PKB models additive effects. It is possible for pathways without strong marginal signals to be picked out by PKB in the presence of other pathways in the model.

\subsection{Comparison with models using only clinical features}

\begin{figure}[htp]
	\centering
	\includegraphics[width = 0.95\textwidth]{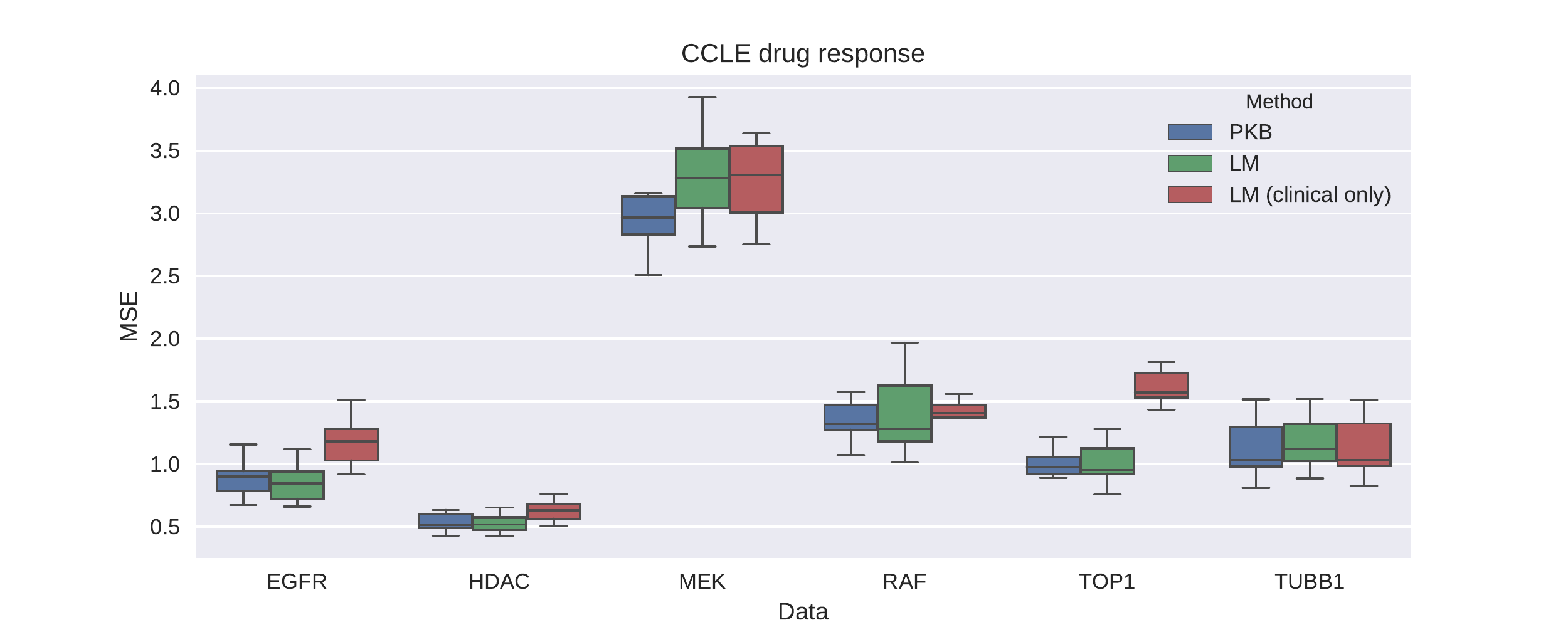} \\ \vspace{-4mm}
	\includegraphics[width = 0.95\textwidth]{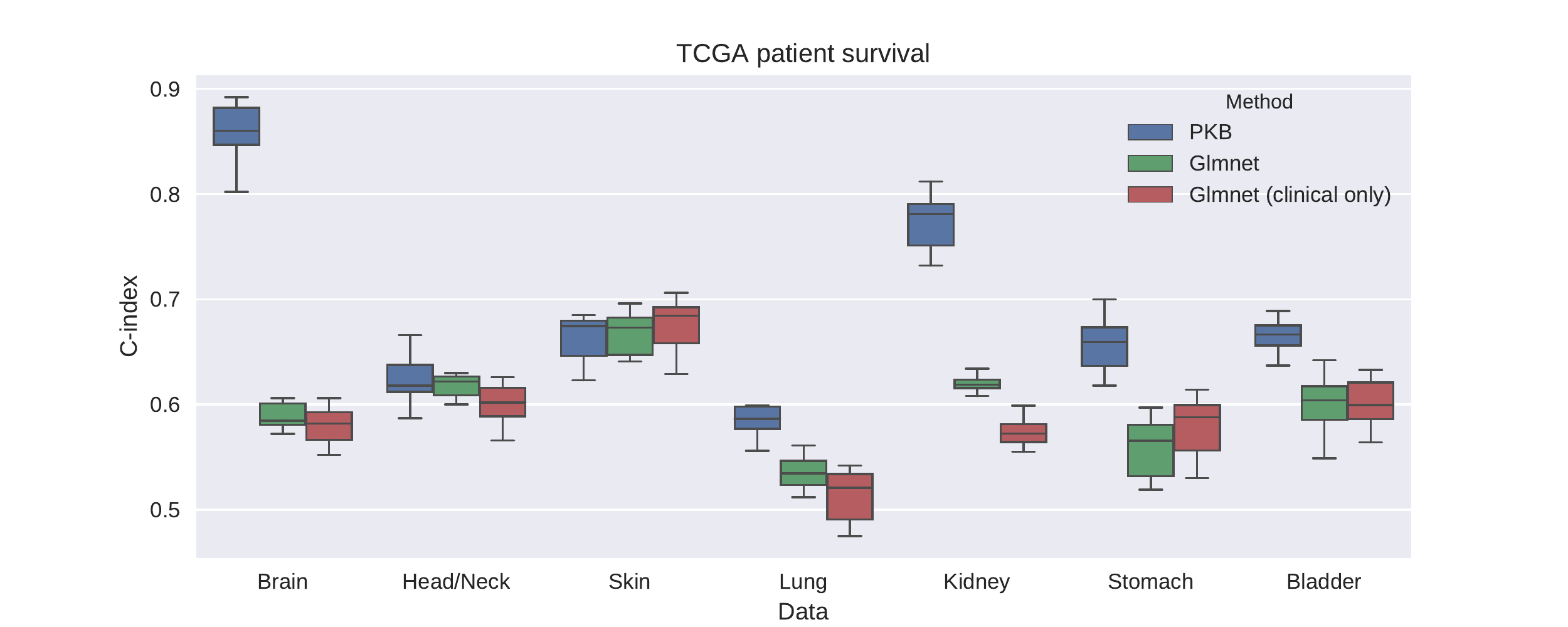}
	\caption{Prediction performances from models with and without genomic features. The figure presents the prediction accuracy from three types of models on each dataset. The models include: PKB, linear model, and linear model with only clinical features as predictors. LM in the upper panel represents linear regression model, which reports the best results from LASSO, Ridge Regression, and ElasticNet. The methods with label ``clinical only" are trained without genomic features. The PKB boxes represent the best results from PKB-$L_1$ and PKB-$L_2$.}
	\label{fig:clin_noclin}
\end{figure}

The PKB score function contains a linear model of clinical features and a nonlinear model of genomic features. It is natural to ask how much improvement the genomic part brings to the prediction accuracy, compared to the clinical part. Considering that genomic data is much more expensive to acquire compared to clinical data, it makes sense to acquire genomic data only when the prediction accuracy can indeed be improved. We compared the performances of the following three methods in both the CCLE and TCGA datasets: the PKB methods which utilize clinical, genomic, and pathway information; linear models with both clinical and genomic features; and linear models with only clinical features. The results are presented in \textbf{Figure \ref{fig:clin_noclin}}. 

The upper panel of \textbf{Figure \ref{fig:clin_noclin}} shows the MSEs of the three methods on CCLE drug response datasets. The ``LM" method represents the best results from LASSO, Ridge Regression, and ElasticNet. In four datasets (EGFR, HDAC, MEK, and TOP1), using genomic information significantly improved the prediction MSE. In the MEK dataset, the MSE difference between the linear models with and without genomic features is not significant. However, using PKB can improve the accuracy significantly.

Comparison of the three models in the TCGA datasets is presented in the lower panel of \textbf{Figure \ref{fig:clin_noclin}}. Glmnet was used to fit the linear models. Only in the skin cancer dataset, using genomic information failed to offer predictive signals in addition to the clinical features. In the six other datasets, PKB achieved significantly higher C-index than the clinical-only Glmnet. However, when the genomic features were modeled using Glmnet, the gain in C-index became moderate, and even failed to outperform the clinical-only version on the stomach and bladder cancer datasets. The results indicate that PKB is able to capture genomic signals more efficiently than Glmnet.

\section{Discussion}
In this article, we have extended the PKB framework proposed by  Zeng et al.\cite{zeng2019pathway} to perform regression and survival analysis and incorporate clinical features as predictors by adding a linear part to the base learner spaces. We have applied PKB to the CCLE datasets to predict cancer drug responses on cell lines, and to the TCGA datasets to predict cancer patients' survival. In both applications, PKB achieved equal or superior prediction accuracy than competing methods in most datasets. Especially in several TCGA datasets, PKB had significantly improved the prediction accuracy. We have further compared PKB with linear models that only use clinical predictors. These results indicate that PKB can effectively capture genomic predictive signals in addition to clinical signals, and significantly improve prediction performance.

In the PKB regression model, the final score $F(\bd{x}, \bd{z})$ function is an estimate of the regression function, which can be used directly to predict the outcome variable. However, in the survival model, $F(\bd{x}, \bd{z})$ is not an estimate of the patient's survival time, but rather an estimate to the risk score in the hazard function (\textbf{Equation (\ref{eqn:hazard})}). Since the baseline hazard $h_0(t)$ is still unknown, we cannot provide direct estimates of patients' survival times. Nonetheless, after an estimate for $F(\bd{x}, \bd{z})$ is acquired, it is easy to nonparametrically estimate the baseline hazard function, and subsequently estimate the survival times.

We have also assumed that the clinical effects and pathway effects are additive, and therefore no interactions between clinical features and pathways are modeled. In the presence of categorical clinical features, informative pathways are supposed to be informative in all categories. This assumption, however, may be violated in real datasets. For example, each CCLE dataset is a mixture of cell lines from several cancer types. Even if a pathway is significant in one or two cancer types, it might not be predictive for all cancer types. \cite{sanchez2018oncogenic} This may be why PKB has not made significant improvement in certain CCLE prediction tasks. We have also tried to fit PKB for each cancer type separately, but the small sample sizes made it difficult to identify true pathway signals. 

In the calculation of kernel functions, all the genes in the same pathways have been treated equally. It is possible that, by giving larger weights to important genes, the PKB models can achieve better prediction accuracy. Suppose gene $i$ takes weight $w_i > 0$. We can modify the calculation of the kernel functions to focus more on the highly weighted genes and we propose the following modifications to the radial basis kernel and polynomial kernel functions, respectively:
$$ K_{\text{rbf}}(\bd{u}, \bd{v})  =  \exp \left[ - \frac{\sum_j w_j (u_j - v_j)^2}{\sum_j w_j}  \right], \quad
K_{\text{poly}}(\bd{u}, \bd{v})  =  \left( 1 + \frac{ \sum_j w_j u_j v_j }{ \sum_j w_j } \right)^d.$$
We have explored this idea using gene weights calculated from GeneMANIA, \citep{warde2010genemania} where we can acquire physical interaction network between genes. The edges between genes are annotated with interaction strength, and the total degrees of the genes are used as the weights. We have utilized the above weighted kernel function to fit PKB models. However, we did not observe significant difference in prediction performance compared to the unweighted version. Detailed results are available in \textbf{Appendix G} in the \textbf{Supplementary Materials}. We leave gene weights as an optional parameter when using PKB, so that users are able to explore different ways to calculate weights for improved prediction accuracy.

The current PKB algorithm can also be improved for better computational efficiency. The 3-fold cross-validation step is the most time and space-consuming part of the model, since it involves running three boosting processes at the same time. It is possible to adopt the notion of out-of-bag (OOB) samples from RandomForest and GBR to calculate testing loss in just one boosting process. In each iteration, instead of using all the samples, we draw a bootstrap sample to train the increment function. The samples not selected in the training set are treated as OOB samples, on which testing loss can be computed. It has been reported that OOB often underestimates the optimal number of iterations, \citep{ridgeway2006gbm} but brings the advantage of efficient model training, especially when the dataset is large.

A Python software implementing the PKB algorithms is available from Github repository: \href{https://github.com/zengliX/PKB}{https://github.com/zengliX/PKB2}. The data that support the findings of this study are available from the corresponding author upon reasonable request.

\section*{Acknowledgment}
This work was supported in part by NIH grants P50 CA196530 and P30 CA016359.

	\newpage
\newpage
\begin{figure}[h]
	\centering
	\includegraphics[width=\textwidth]{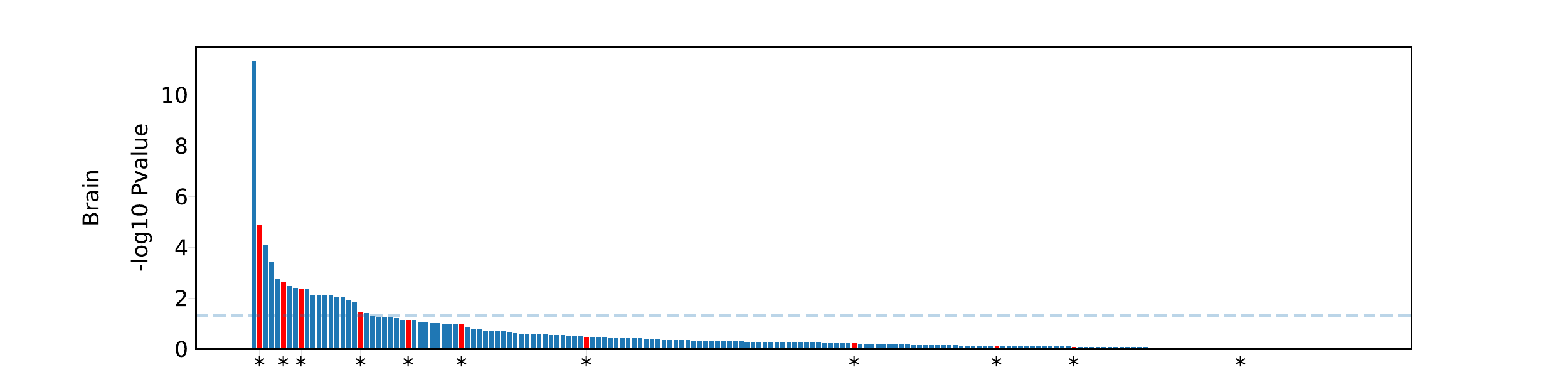}\\ \vspace{-3mm}
	\includegraphics[width=\textwidth]{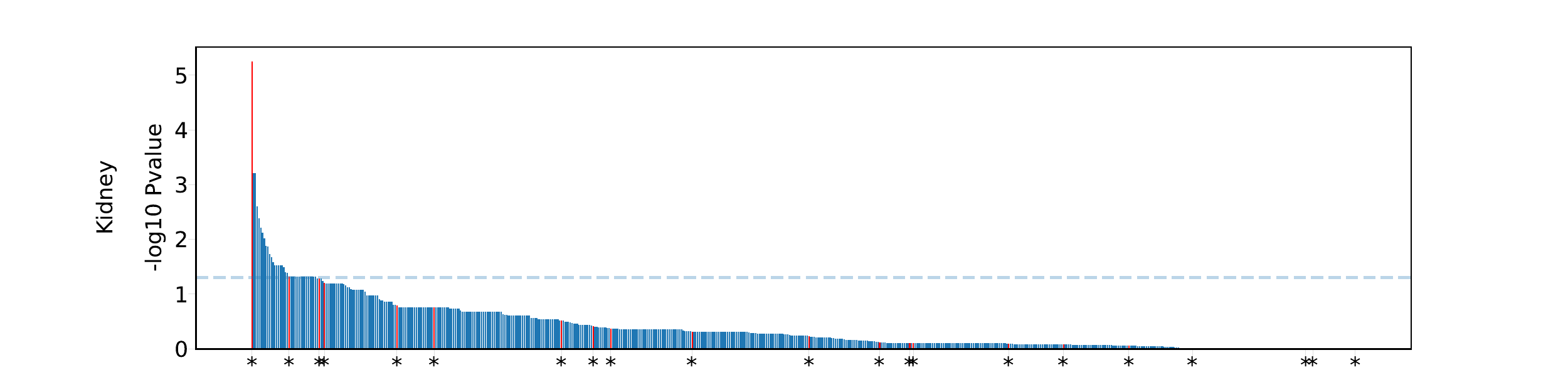}\\ \vspace{-3mm}
	\includegraphics[width=\textwidth]{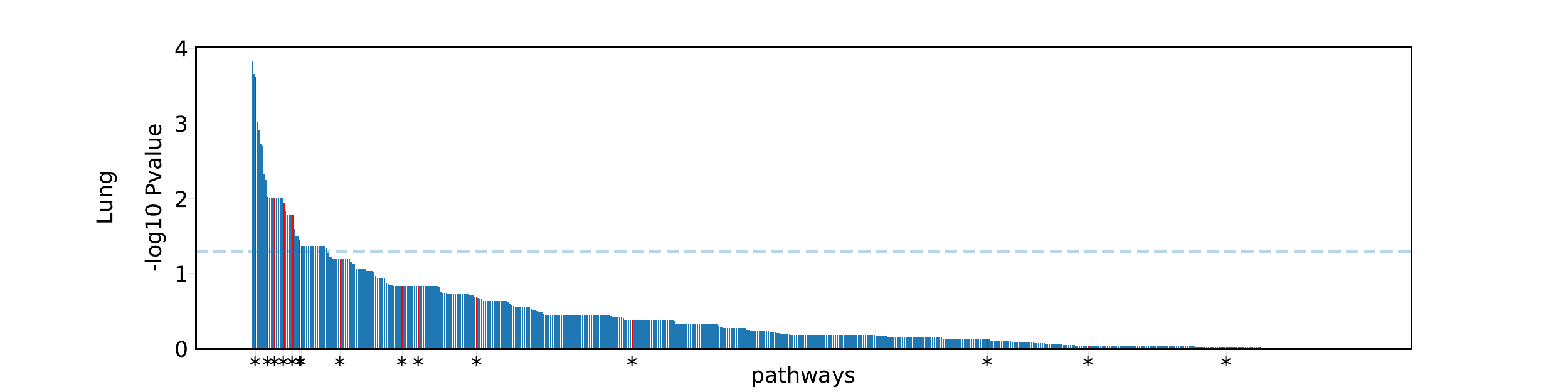}
	\caption{Enrichment analysis on brain, kidney, and lung cancer datasets. X-axis represents pathways sorted by their p-values in the enrichment analysis. The blue dashed line corresponds to p-value 0.05. The pathways marked with red bars and stars are pathways with significant weights in PKB.}
	\label{fig:gsea}
\end{figure}
\newpage
	\newpage
	\begin{table}[htp]
		\centering
		\begin{tabular}{lllllllll}
			\hline
			\multirow{2}{*}{Method} & \multicolumn{2}{c}{Model 1}     &  & \multicolumn{2}{c}{Model 2}     &  & \multicolumn{2}{c}{Model 3}   \\ \cline{2-3} \cline{5-6} \cline{8-9} 
			& M = 20         & M = 50         &  & M = 20         & M = 50         &  & M = 20        & M = 50        \\ \hline
			PKB-$L_1$                  & \textbf{17.11} & \textbf{22.82} &  & \textbf{32.92} & \textbf{33.94} &  & \textbf{7.22} & \textbf{8.37} \\
			PKB-$L_2$                  & \textbf{16.84} & \textbf{22.41} &  & \textbf{31.25} & \textbf{33.65} &  & \textbf{5.4}  & \textbf{5.77} \\
			LASSO                   & 34.85          & 40.58          &  & 49.74          & 50.94          &  & 21.15         & 22.88         \\
			Ridge                   & 50.71          & 53.1           &  & 57.32          & 60.52          &  & 23.25         & 25.99         \\
			ElasticNet              & 34.87          & 42.21          &  & 49.27          & 50.93          &  & 21.25         & 23.51         \\
			RandomForest            & 46.88          & 50.94          &  & 55.34          & 57.34          &  & 21.09         & 22.67         \\
			GBR                     & 48.54          & 51.75          &  & 50.9           & 52.82          &  & 21.46         & 23.7          \\
			SVR                     & 50.02          & 53.58          &  & 56.12          & 59.58          &  & 19.06         & 24.31         \\ \hline
		\end{tabular}
		\caption{Cross-validated MSE of all methods on simulated regression datasets. $M$ represents the number of simulated pathways. For each dataset, MSE values from the top two methods are in boldface.}
		\label{tab:simu_reg}
	\end{table}

\newpage
\begin{table}[htp]
	\centering
	\begin{tabular}{lllllllll}
		\hline
		\multirow{2}{*}{Method} & \multicolumn{2}{c}{Model 1}  & \multicolumn{1}{c}{} & \multicolumn{2}{c}{Model 2}   & \multicolumn{1}{c}{} & \multicolumn{2}{c}{Model 3}   \\ \cline{2-9} 
		& M = 20       & M = 50        &                      & M = 20        & M = 50        &                      & M = 20        & M = 50        \\ \hline
		PKB-$L_1$                  & \textbf{0.9} & \textbf{0.86} & \textbf{}            & \textbf{0.77} & \textbf{0.77} & \textbf{}            & \textbf{0.88} & \textbf{0.87} \\
		PKB-$L_2$                 & \textbf{0.9} & \textbf{0.88} & \textbf{}            & \textbf{0.72} & \textbf{0.76} & \textbf{}            & \textbf{0.9}  & \textbf{0.89} \\
		Glmnet                  & 0.78         & 0.79          &                      & 0.69          & 0.71          &                      & 0.65          & 0.66          \\
		RandomSurvivalForest    & 0.67         & 0.67          &                      & 0.65          & 0.67          &                      & 0.63          & 0.64          \\
		CoxBoost                & 0.78         & 0.78          &                      & 0.7          & 0.7          &                      & 0.66          & 0.66          \\ \hline
	\end{tabular}
	\caption{Cross-validated C-indices of all methods on simulated survival datasets. $M$ represents the number of simulated pathways. For each dataset, C-indices from the top two methods are highlighted in boldface.}
	\label{tab:simu_surv}
\end{table}

\newpage

\begin{table}[htp]
	\centering
	\begin{tabular}{lllllll}
		\hline
		Method       & EGFR          & HDAC          & MEK           & RAF           & TOP1          & TUBB1         \\ \hline
		PKB-$L_1$       & 0.88          & \textbf{0.54} & \textbf{2.92} & \textbf{1.37} & \textbf{0.98} & \textbf{1.12} \\
		PKB-$L_2$       & 0.88          & \textbf{0.54} & \textbf{2.9}  & \textbf{1.35} & 1.01          & \textbf{1.12} \\
		LASSO        & \textbf{0.87} & 0.62          & 3.63          & 1.47          & 1.14          & 1.18          \\
		Ridge        & \textbf{0.84} & \textbf{0.52} & 3.28          & 1.39          & 1.01          & 1.19          \\
		ElasticNet   & 0.89          & 0.59          & 3.37          & 1.44          & 1.14          & 1.17          \\
		RandomForest & 0.88          & 0.56          & 3.17          & \textbf{1.37} & \textbf{0.99} & \textbf{1.12} \\
		GBR          & 0.91          & \textbf{0.54} & 3.3           & 1.43          & 1.0           & 1.14          \\
		SVR          & 0.88          & 0.55          & 3.17          & 1.38          & \textbf{0.99} & \textbf{1.11} \\ \hline
	\end{tabular}
	\caption{Cross-validated MSEs from all methods on CCLE drug response data. For each dataset, the two methods with the smallest MSEs are marked in boldface.}
	\label{tab:ccle_mean}
\end{table}

\newpage

\begin{table}[htp]
	\centering
	\begin{tabular}{llllllll}
		\hline
		Method               & Brain         & Head/Neck          & Skin      & Lung    & Kidney        & Stomach       & Bladder       \\ \hline
		PKB-$L_1$               & \textbf{0.86} & \textbf{0.62} & 0.66          & \textbf{0.59} & \textbf{0.78} & \textbf{0.66} & \textbf{0.67} \\
		PKB-$L_2$               & \textbf{0.85} & 0.61          & 0.64          & \textbf{0.57} & \textbf{0.77} & \textbf{0.65} & \textbf{0.67} \\
		Glmnet               & 0.59          & \textbf{0.62} & \textbf{0.67} & 0.54          & 0.62          & 0.56          & 0.6           \\
		RandomSurvivalForest & 0.58          & 0.54          & 0.63          & 0.54          & 0.58          & 0.54          & 0.52          \\
		CoxBoost             & 0.59          & \textbf{0.62} & \textbf{0.67} & 0.53          & 0.61          & 0.53          & 0.6           \\ \hline
	\end{tabular}
	\caption{Cross-validated C-index from all methods on CCLE drug response data. For each dataset, the two methods with the highest C-index are marked in boldface.}
	\label{tab:tcga_mean}
\end{table}
\newpage
\section*{Appendix A. Calculation of gradient and Hessian matrix} \label{sec:calcu}
In this section, we provide closed form expressions for calculating the gradients $\grad{L}{F}$ and Hessian matrices $\hess{F}$ for classification and survival model, as needed in Section 2.1 in the main text.

\noindent $\bullet$ Classification model:
\begin{eqnarray*}
	(\grad{L}{F})_i & = & \frac{\partial L(\bd{y}, \bd{F})}{\partial F(\bd{x}_i, \bd{z}_i)} = -\frac{y_i}{1+\exp[y_i F(\bd{x}_i, \bd{z}_i)]} \\
	(\hess{F})_{ij} & = & \frac{\partial^2 L(\bd{y}, \bd{F})}{\partial F(\bd{x}_i, \bd{z}_i) \partial F(\bd{x}_j, \bd{z}_j)} \\
	& = & \begin{cases}
		\frac{\exp[y_iF(\bd{x}_i, \bd{z}_i]}{(1+\exp[y_iF(\bd{x}_i, \bd{z}_i)])^2} &\text{if  $i = j$}\\
		0 &\text{if $i \neq j$}
	\end{cases}
\end{eqnarray*}
\noindent $\bullet$ Survival model:
\begin{eqnarray*}
	(\grad{L}{F})_i &=&   \frac{\partial L(\bd{y}, \bd{F})}{\partial F(\bd{x}_i, \bd{z}_i)}  = -\frac{1}{N}\left( \delta_i - \sum^N_{j=1}\delta_j1_{\left\lbrace t_i \geq t_j\right\rbrace }\frac{\exp[F(\bd{x}_i, \bd{z}_i)]}{\sum^N_{r=1}1_{\left\lbrace t_r \geq t_j \right\rbrace }\exp[F(\bd{x}_r, \bd{z}_r)]}\right) \\
	(\hess{F})_{ij} & = & \frac{\partial^2 L(\bd{y}, \bd{F})}{\partial F(\bd{x}_i, \bd{z}_i) \partial F(\bd{x}_j, \bd{z}_j)} \\
	& = &  \frac{1}{N} \sum^N_{l=1}\delta_j 1_{\left\lbrace t_i \geq t_l\right\rbrace }  \frac{1_{\{ i = j\}}\exp[F(\bd{x}_i, \bd{z}_i)]}{ \sum^N_{r=1}1_{\left\lbrace t_r \geq t_l \right\rbrace }\exp[F(\bd{x}_r, \bd{z}_r)]} - \\
	& & \frac{1}{N} \sum^N_{l=1}\delta_j 1_{\left\lbrace t_i \geq t_l\right\rbrace } 1_{\left\lbrace t_j \geq t_l \right\rbrace }\frac{\exp[F(\bd{x}_i, \bd{z}_i)]\exp[F(\bd{x}_j, \bd{z}_j)]}{\left( \sum^N_{r=1}1_{\left\lbrace t_r \geq t_l \right\rbrace }\exp[F(\bd{x}_r, \bd{z}_r)]\right) ^2} 
\end{eqnarray*}
Since the regression model's loss function is quadratic itself, we do not need to calculate the gradient or Hessian matrix to make approximation. As a result of using the negative log partial likelihood as loss in survival model, the loss for each sample is dependent on the $F(\bd{x}, \bd{z})$ values of other samples, which leads to non-zero off-diagonal values in the Hessian matrix. It is more complex compared to the diagonal Hessian matrix in classification problem, and therefore causing the algorithm to be slower than classification as well.
\newpage
\section*{Appendix B. Proof of the reduction to LASSO and Ridge Regression} \label{sec:proof}
In this section, we prove the following result on Equation (3) as stated in Section 2.1:

By applying transformation
\begin{eqnarray*}
	\tilde{\eta} &=& \frac{1}{\sqrt{2}}\hess{F}^{\frac{1}{2}}(I_N - Z(Z^T \hess{F} Z)^{-1}Z^T \hess{F})\hess{F}^{-1}\grad{L}{F}\\
	\tilde{K}_m &=& \frac{1}{\sqrt{2}}\hess{F}^{\frac{1}{2}}(I_N - Z(Z^T\hess{F}Z)^{-1}Z^T\hess{F})K_m,
\end{eqnarray*}
solving $\beta$ in Equation (3) is reduced to a penalized linear regression problem
\begin{equation*}
\min_{\beta} \|\tilde{\eta} + \tilde{K}_m \beta \|_2^2 + \lambda \Omega(f).
\end{equation*}

Because the penalty term $\Omega(f)$ in Equation (3) does not involve $\gamma$, we can fix $\beta$ and solve $\gamma$ as a weighted least squares problem:
$$\gamma = -(Z^T\hess{F}Z)^{-1}Z^T\hess{F}(K_m\beta + \hess{F}^{-1}\grad{L}{F})$$
Plugging it back to Equation (3), we can get the following reduction:
\begin{eqnarray*}
	& & \min_{\beta, \gamma} \frac{1}{2} (K_m \beta + Z \gamma + \hess{F}^{-1} \grad{L}{F})^T \hess{F} (K_m \beta + Z \gamma + \hess{F}^{-1} \grad{L}{F}) + \lambda \Omega(f) \\
	& = & \min_{\beta} \frac{1}{2} [ U(K_m \beta + \hess{F}^{-1} \grad{L}{F}) ]^T \hess{F} [ U (K_m \beta + \hess{F}^{-1} \grad{L}{F}) ]  + \lambda \Omega(f) \\
	& = & \min_{\beta}  \| \frac{1}{\sqrt{2}}\hess{F}^{\frac{1}{2}}U(K_m \beta + \hess{F}^{-1} \grad{L}{F}) \|_2^2+ \lambda \Omega(f) \\
	&=& \min_{\beta} \|\tilde{\eta} + \tilde{K}_m \beta \|_2^2 + \lambda \Omega(f), 
\end{eqnarray*}
which proves our statement. In the equations, the matrix $U = (I_N -Z(Z^T\hess{F}Z)^{-1}Z^T\hess{F})$, is an intermediate matrix.

\newpage
\section*{Appendix C. Calculation of C-index}
\label{sec:Cind}
In this section, we describe the calculation of C-index for evaluation of survival prediction accuracy. For each pair of samples, assume they have survival outcomes $(t_i, \delta_i)$ and $(t_j, \delta_j)$, and estimated risk scores $f_i$ and $f_j$. The concordance for the pair is calculated as follows:
\begin{itemize}
	\item If the sample with smaller $t$ is censored, omit the pair
	\item If $t_i = t_j$ and both samples are censored, omit the pair
	\item If $t_i \neq t_j$\\
	\qquad - If  $f_i = f_j$, concordance equals to 0.5\\
	\qquad - If the sample with higher risk has smaller $t$, then concordance equals to 1, otherwise 0
	\item If $t_i = t_j$\\
	\qquad - If $\delta_i = \delta_j = 1$, concordance equals to 1 if $f_i = f_j$, otherwise 0.5\\
	\qquad - If the censored sample has smaller risk score, concordance is 1, otherwise 0
	
\end{itemize}
We look at all pairs of samples, and the C-index is defined as
$$\text{C-index} = \frac{ \text{Total Concordance} }{\# \text{Permissible pairs} }$$
\newpage
\section*{Appendix D. Parameter tuning}
\label{sec:tune}
The usage of most methods in this article involves tuning of certain model parameters. In general, we generated a series of candidate values for each parameter, and considered all possible combinations among them. Cross-validation was used to determine the optimal combination and corresponding prediction accuracy.

\begin{itemize}
	\item PKB\\
	In simulations, we used the following candidate values for tuning parameters:\\
	1. kernel function: radial basis function (rbf), polynomial kernel with degree 3 (poly3)\\
	2. learning rate: 0.01, 0.05\\
	3. penalty multiplier: 0.04, 0.2, 1\\
	An automatic procedure is proposed in Zeng et al\cite{zeng2019pathway} to determine the penalty parameter, but it is often overly strong. Therefore, we tried the penalty multipliers above ($\leq 1$) to soften the penalty term. In real data applications, we used the following candidate values:\\
	1. kernel function: rbf, poly3\\
	2. learning rate: 0.005, 0.03\\
	3. penalty multiplier: 0.04, 0.2, 1\\
	4. pathway databases: KEGG, Biocarta, GO-BP
	
	\item LASSO/ Ridge Regression/ Elastic Net\\
	The LASSO solver in python \ttt{sklearn} module has a default method for generating a series of feasible penalty parameters $\lambda$. We used it to generate 100 $
	\lambda$s and evaluated prediction performances through cross-validation. The optimal value was used for prediction.\\
	For Ridge Regression, we used 100 values between 0.1 and 1,000 (equally spaced on log scale) as candidate penalty parameters. The value demonstrating the optimal prediction performance in cross-validation was used to predict test data.\\
	The Elastic Net method has penalty term $ r \alpha \|\beta\|_1 + (1-r)\alpha/2 \|\beta\|^2_2$, which involves two tuning parameters, $r$ and $\alpha$. $\alpha$ can be considered as the total penalty, and $r$ determines how much of it is assigned to the $L_1$ part. In our parameter tuning, we considered all combinations of $r = 0.1, 0.2, \ldots, 1$ and 30 $\alpha$ values generated by default.
	
	\item RandomForest\\
	We tuned the parameters, number of trees and maximum depth for a single tree, for optimized prediction performance. We had a list of candidate values for each parameter, and fit prediction models using all possible parameter value combinations. Results for the optimal combination were reported. The candidate values we used for both parameters are as follows:\\
	1. number of trees in forest: 500 and 800\\
	2. maximum tree depth: 3, 5, 7, and no constraint
	\item Gradient Boosting Regression\\
	The tuning parameters are: maximum tree depth and learning rate. The tuning process is similar to the procedure in RandomForest. For each parameter combination, we used out-of-bag (OOB) samples (1/3 of training data in each iteration) to decide optimal number of iterations, and used the optimal iteration number to fit a model without OOB sample. The candidate values for tuning parameters are:\\
	1. maximum tree depth: 3, 5, and 7\\
	2. learning rate: 0.002, 0.01, and 0.05
	\item SVR\\
	We tuned three parameters of SVR: $C$, which adjusts the penalty on samples violating the $\epsilon$-tube; $\epsilon$, which is the width of the $\epsilon$-tube; and the kernel functions. The candidate values for these parameters are:\\
	1. $C$: 10 values equally spaced on log scale between 0.1 and 1000\\
	2. $\epsilon$: 10 values equally spaced between 0.1 and 4\\
	3. kernel functions: rbf, poly2, and poly3
	
	\item Glmnet\\
	The parameters we tuned for Glmnet included:  $\alpha \in (0, 1)$, which is the ElasticNet mixing parameter, determining the fraction of $L_1$ penalty; $\lambda$, which adjusts the overall strength of the penalty. Candidate values for them are:\\
	1.	$\alpha$: 20 values equally spaced between 0 and 1\\
	2.	$\lambda$: 0.001, 0.01, 0.1, 1, 2, 5, and 10
	\item RandomSurvivalForest\\
	We tuned the parameters including the splitting rule of the trees, the number of trees, and the depth of the trees. Candidate values for the parameters are:\\
	1.	Splitting rules: logrank (splitting nodes by maximization of the log-rank test statistic), and logrankscore (splitting nodes by maximization of the standardized log-rank statistic)\\
	2.	The number of trees: 100, 200, and 500\\
	3.	The maximum tree depth: 3, 5, and 7
	
	\item CoxBoost\\
	The tuning parameters for CoxBoost were step-size factor, which determined the step-size modification factor by which each boosting step size should be changed; and the number of iterations, which controlled the total number of boosting steps. \\
	1.	Step-size factor: 0.01, 0.1, 0.5, 1, and 2\\
	2.	The number of boosting iterations: 100, 250, and 500 
\end{itemize}
\newpage
\section*{Appendix E. Pathway enrichment analysis in TCGA survival datasets}\label{sec:gsea}
The pathway enrichment analysis was performed based on the following steps:
\begin{itemize}
	\item[1.] For each gene, we performed a Cox regression using the clinical features and the gene's expression as predictors. P-values for each gene were calculated from the Cox models.
	\item[2.] We chose $20\%$ genes with the smallest p-values as relevant genes to patients' survival times. 
	\item[3.] The relevant genes were used to perform Fisher's exact test-based enrichment analysis \cite{huang2008bioinformatics} on the pathway database (KEGG, GO-BP, or Biocarta) used in the PKB model that achieved the highest accuracy. An enrichment p-value was calculated for each pathway.
\end{itemize}

\begin{figure}[htp]
	\centering
	\includegraphics[width=\textwidth]{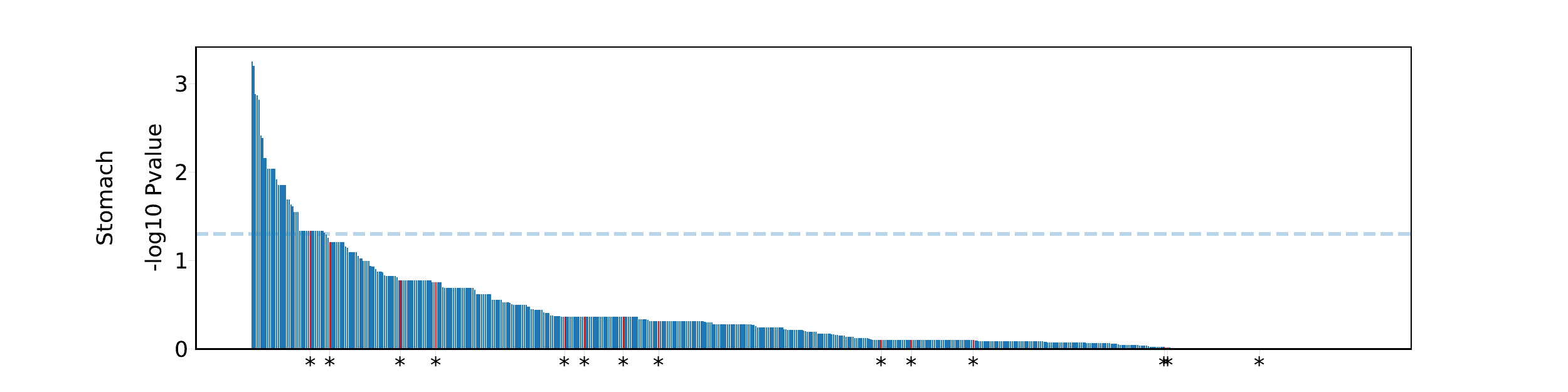}\\ \vspace{-3mm}
	\includegraphics[width=\textwidth]{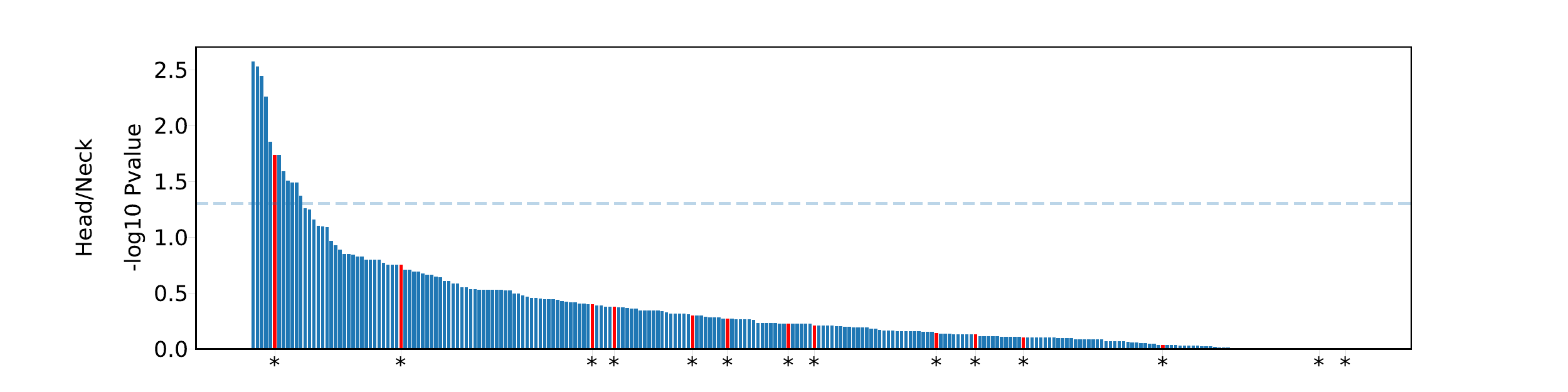}\\ \vspace{-3mm}
	\includegraphics[width=\textwidth]{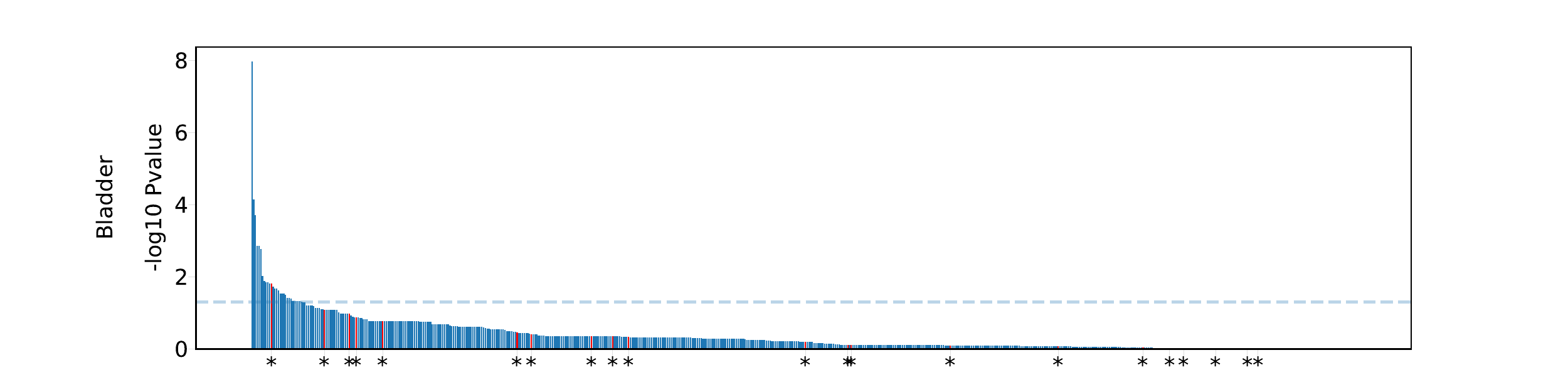}\\ \vspace{-3mm}
	\includegraphics[width=\textwidth]{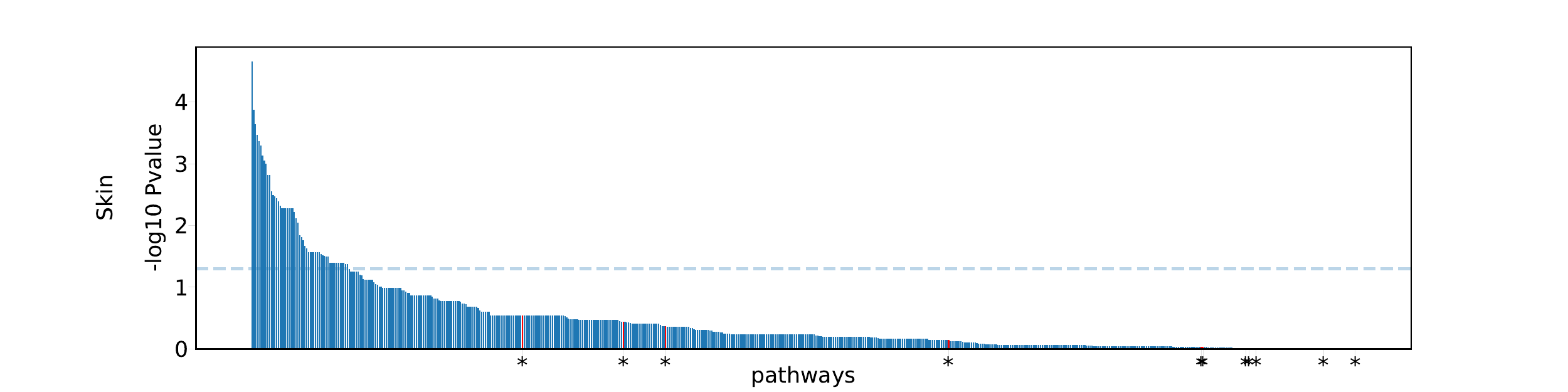}
	\caption{Enrichment analysis on stomach, head/neck, bladder, and skin cancer datasets. X-axis represents pathways sorted by their p-values in the enrichment analysis. The blue dashed line corresponds to p-value 0.05. The pathways marked with red bars and stars are pathways with significant weights in PKB.}
\end{figure}
\newpage
\section*{Appendix F. PKB model pathway weights from survival simulations}
\label{sec:surv_weights}
\begin{figure}[htp]
	\centering
	\includegraphics[width=1\textwidth]{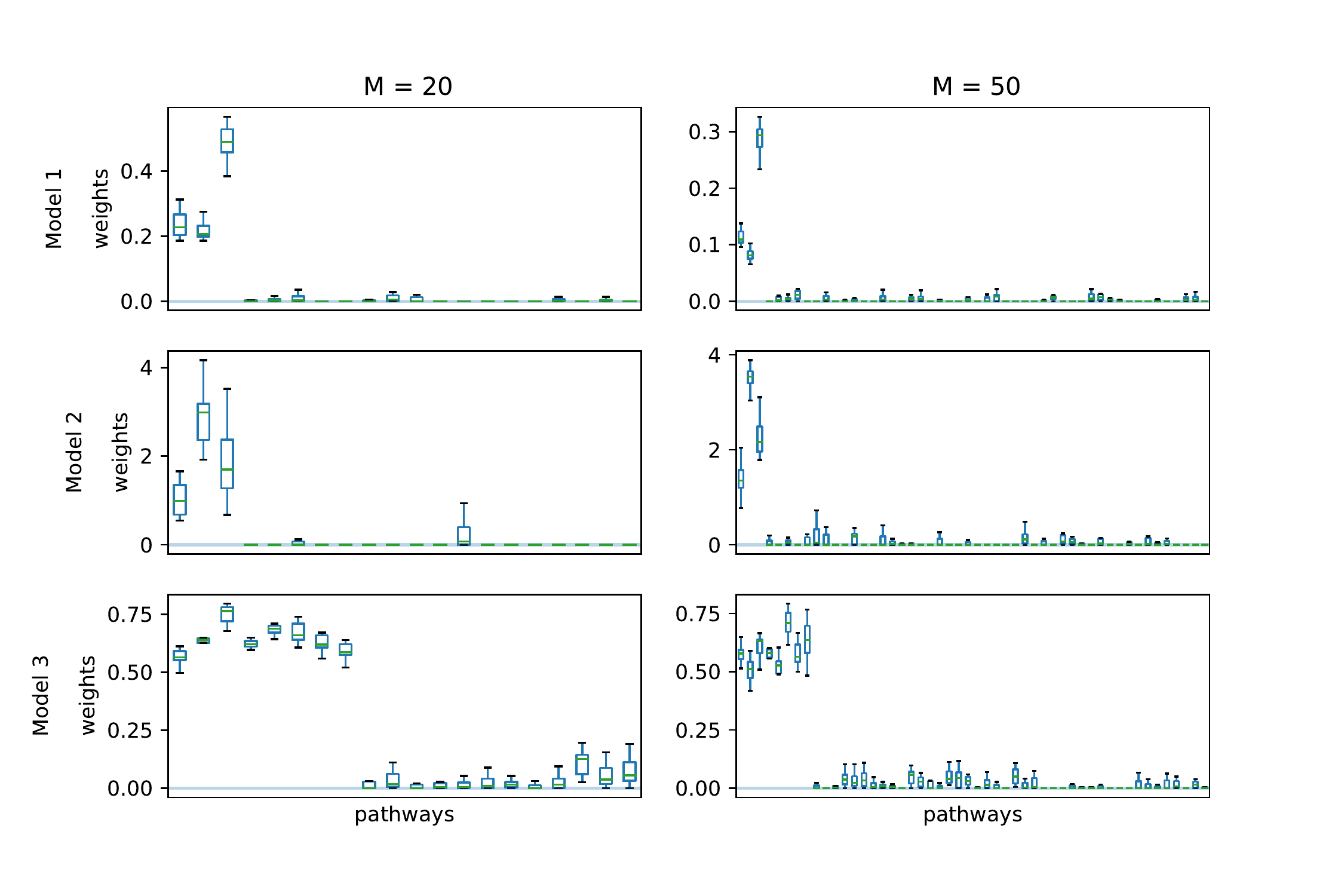}
	\caption{Boxplots for pathway weights in the survival simulations. Each box represents the weights distribution for one pathway over ten PKB runs.}
\end{figure}
\newpage
\section*{Appendix G. Weighted PKB prediction results}
\label{sec:weightedPKB}

\begin{figure}[htp]
	\centering
	\includegraphics[width=1\textwidth]{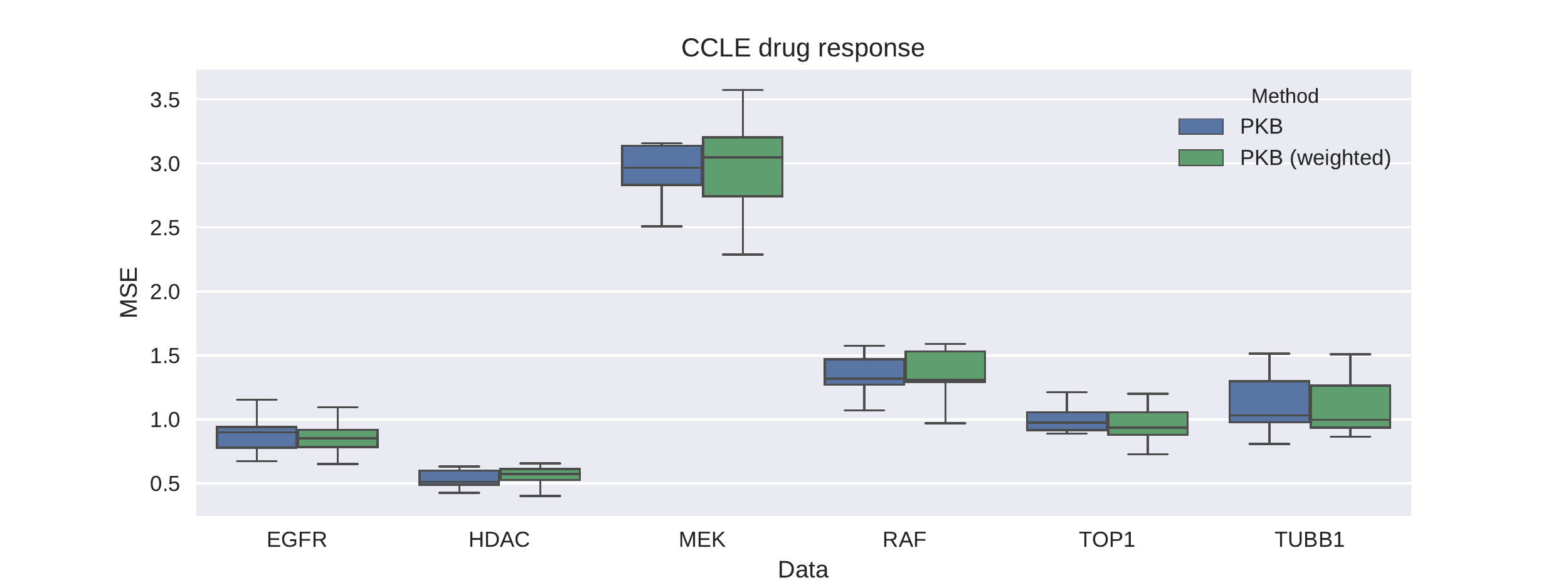}\\ \vspace{-3mm}
	\includegraphics[width=1\textwidth]{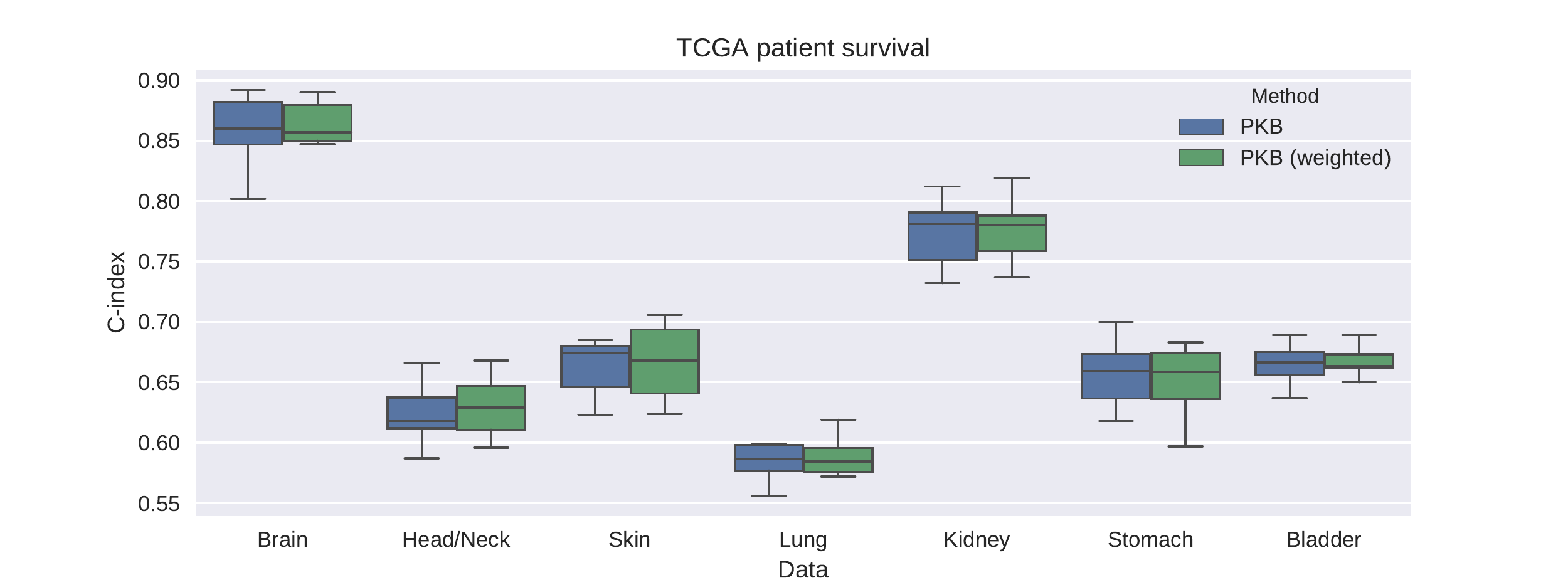}
	\caption{Prediction performances from PKB and PKB with gene weights. The upper panel demonstrates prediction MSEs from CCLE drug response datasets. The lower panel demonstrates prediction C-index from TCGA patient survival datasets. No significant difference in accuracy can be observed from the two methods.}
\end{figure}
\newpage
\section*{Appendix H. Standard deviations for the prediction accuracy measures in simulations and real data appications}
\label{sec:std}

% \usepackage{multirow}
\begin{table}[htp]
	\centering
	\begin{tabular}{lllllllll}
		\hline
		\multirow{2}{*}{Method} & \multicolumn{2}{c}{Model 1} &  & \multicolumn{2}{c}{Model 2} &  & \multicolumn{2}{c}{Model 3} \\ \cline{2-3} \cline{5-6} \cline{8-9} 
		& M = 20       & M = 50       &  & M = 20       & M = 50       &  & M = 20       & M = 50       \\ \hline
		PKB-$L_1$                  & 2.02         & 7.63         &  & 5.73         & 4.26         &  & 0.95         & 0.77         \\
		PKB-$L_2$                & 2.09         & 6.67         &  & 5.34         & 3.93         &  & 0.59         & 0.58         \\
		LASSO                   & 4.21         & 11.35        &  & 5.91         & 8.84         &  & 1.81         & 2.2          \\
		Ridge                   & 3.81         & 13.03        &  & 8.46         & 9.42         &  & 1.81         & 2.19         \\
		ElasticNet              & 4.17         & 10.64        &  & 6.08         & 8.4          &  & 2.05         & 2.43         \\
		RandomForest            & 4.2          & 14.51        &  & 7.38         & 10.43        &  & 1.54         & 2.03         \\
		GBR                     & 4.96         & 12.53        &  & 8.08         & 10.21        &  & 1.72         & 1.8          \\
		SVR                     & 3.61         & 13.16        &  & 8.08         & 9.46         &  & 1.28         & 1.84         \\ \hline
	\end{tabular}
	\caption{Standard deviations for prediction MSE in regression simulation studies.}
\end{table}

% \usepackage{multirow}
\begin{table}[htp]
	\centering
	\begin{tabular}{lllllllll}
		\hline
		\multirow{2}{*}{Method} & \multicolumn{2}{c}{Model 1} & \multicolumn{1}{c}{} & \multicolumn{2}{c}{Model 2} & \multicolumn{1}{c}{} & \multicolumn{2}{c}{Model 3} \\ \cline{2-3} \cline{5-6} \cline{8-9} 
		& M = 20       & M = 50       &                      & M = 20       & M = 50       &                      & M = 20       & M = 50       \\ \hline
		PKB-$L_1$                  & 0.01         & 0.02         &                      & 0.03         & 0.03         &                      & 0.01         & 0.01         \\
		PKB-$L_2$                  & 0.01         & 0.02         &                      & 0.04         & 0.05         &                      & 0.01         & 0.01         \\
		Glmnet                  & 0.02         & 0.02         &                      & 0.02         & 0.01         &                      & 0.02         & 0.01         \\
		RandomSurvivalForest    & 0.03         & 0.03         &                      & 0.03         & 0.03         &                      & 0.03         & 0.03         \\
		CoxBoost                & 0.02         & 0.02         &                      & 0.03         & 0.02         &                      & 0.02         & 0.02         \\ \hline
	\end{tabular}
	\caption{Standard deviations for prediction C-index in survival simulation studies.}
\end{table}

\newpage

\begin{table}[htp]
	\centering
	\begin{tabular}{lllllll}
		\hline
		Method       & EGFR & HDAC & MEK  & RAF  & TOP1 & TUBB1 \\ \hline
		PKB-$L_1$       & 0.14 & 0.07 & 0.27 & 0.14 & 0.13 & 0.23  \\
		PKB-$L_2$       & 0.14 & 0.07 & 0.28 & 0.14 & 0.14 & 0.22  \\
		LASSO        & 0.2  & 0.08 & 0.62 & 0.2  & 0.13 & 0.21  \\
		Ridge        & 0.14 & 0.07 & 0.35 & 0.29 & 0.15 & 0.23  \\
		ElasticNet   & 0.16 & 0.09 & 0.51 & 0.18 & 0.13 & 0.21  \\
		RandomForest & 0.12 & 0.08 & 0.24 & 0.18 & 0.13 & 0.21  \\
		GBR          & 0.14 & 0.08 & 0.27 & 0.14 & 0.13 & 0.23  \\
		SVR          & 0.13 & 0.08 & 0.31 & 0.14 & 0.13 & 0.23  \\ \hline
	\end{tabular}
	\caption{Standard deviations for prediction MSE in CCLE drug response datasets.}
\end{table}

\begin{table}[htp]
	\centering
	\begin{tabular}{llllllll}
		\hline
		Method               & Brain & Head/Neck & Skin & Lung & Kidney & Stomach & Bladder \\ \hline
		PKB-$L_1$               & 0.03  & 0.02 & 0.02     & 0.03 & 0.03   & 0.03    & 0.02    \\
		PKB-$L_2$               & 0.03  & 0.03 & 0.03     & 0.03 & 0.02   & 0.02    & 0.02    \\
		Glmnet               & 0.02  & 0.02 & 0.02     & 0.01 & 0.01   & 0.03    & 0.03    \\
		RandomSurvivalForest & 0.02  & 0.02 & 0.02     & 0.01 & 0.02   & 0.02    & 0.03    \\
		CoxBoost             & 0.02  & 0.01 & 0.02     & 0.02 & 0.02   & 0.03    & 0.02    \\ \hline
	\end{tabular}
	\caption{Standard deviations for prediction C-index in TCGA patient survival datasets.}
\end{table}
\newpage
\bibliography{mybib_DB}
\end{document}